\documentclass[9pt, twocolumn]{extarticle}
\usepackage[
    left=0.75in,
    right=0.75in,
    top=0.8in,
    bottom=1in
]{geometry}

\usepackage{float}
\usepackage{amsmath,amssymb,amsfonts}
\usepackage{booktabs}
\usepackage{multirow}
\usepackage{graphicx}
\usepackage{subcaption}
\usepackage{algorithm}
\usepackage{algpseudocode}
\usepackage{hyperref}
\usepackage{makecell}

\usepackage{xcolor}
\usepackage{natbib}
\usepackage{enumitem}
\usepackage{microtype}
\usepackage{comment}

\title{
\textbf{AnchorRM: Parallel Intra-Batch State Reconstruction with
Anchor-Based Residual Memory for Temporal Graph Neural Networks}
}

\author{
\textbf{Haroon Gharwi\textsuperscript{1*}  \quad Kai Shu\textsuperscript{2+}\quad Yue Dai\textsuperscript{3*}}\\[3pt]
\textsuperscript{*}Department of Computer Science, Illinois Institute of Technology, Chicago, IL, USA \\
\textsuperscript{+}Department of Computer Science, Emory University, Atlanta, GA, USA\\[3pt]
\textsuperscript{1}\texttt{hgharwi@hawk.illinoistech.edu} \quad
\textsuperscript{2}\texttt{kai.shu@emory.edu} \quad
\textsuperscript{3}\texttt{ydai21@illinoistech.edu} 
}

\date{}

\begin{document}

\maketitle

\pagestyle{plain}

\begin{abstract}
Memory-based temporal graph neural networks (TGN) rely on recurrent node
memories to capture evolving interaction histories. However, temporal mini-batching introduces a trade-off between computational efficiency and
temporal fidelity: when a node appears multiple times within the same batch,
later interactions are evaluated using stale pre-batch memory because the
intermediate memory updates have not yet been committed. This problem becomes
more severe as temporal batch size increases.
To overcome this challenge, We propose \textbf{AnchorRM}, an anchor-based residual memory mechanism that reconstructs event-specific intra-batch node states without executing
sequential recurrent updates. \textbf{AnchorRM} treats the last committed node memory
at the beginning of each batch as a pre-batch fixed anchor and uses causally preceding
same-batch interactions to infer the missing state evolution relative to that
anchor. The learned residual then calibrates 
the anchor to reconstruct an event-specific node memory for each interaction. 
Because each event-specific residual is predicted directly from the
fixed anchor and its causal history, reconstructed states do not depend on
previously reconstructed states and can therefore be computed in parallel
within the batch while preserving temporal causality.
Experiments across four temporal graph datasets and three memory-based backbones demonstrate that AnchorRM achieves an average 6.16× training speedup (up to 10.05×) and 28.67× batch scaling (up to 40×). These results demonstrate that AnchorRM substantially improves the scalability and training efficiency of memory-based temporal graph learning while maintaining competitive predictive performance.

\end{abstract}

\section{Introduction}
\label{sec:introduction}

Temporal graphs model systems in which interactions, relationships, and node
states evolve over time, with applications ranging from social and
communication networks to recommender systems and financial transactions
\cite{kazemi2020survey}. Unlike static graph representations, temporal graph
models must account for both the structure of interactions and their temporal
ordering. A prominent class of temporal graph neural networks (TGN)
therefore maintains evolving node representations that summarize interaction
history \cite{kumar2019jodie,trivedi2019dyrep,rossi2020tgn}. In particular,
memory-based architectures such as Temporal Graph Networks (TGN) associate
each node with a persistent memory that is updated as new interactions are
observed, providing an explicit mechanism for carrying historical information
through the event stream \cite{rossi2020tgn}.


The recurrent nature of node memory, however, creates a tension between
temporal fidelity and efficient batch-parallel execution. Ideally, if a node
\(u\) participates in a sequence of interactions
\(e_1,e_2,\ldots,e_k\), the memory used at \(e_k\) should reflect the
effects of all causally preceding interactions involving \(u\). Such
sequential state propagation preserves the temporal dependency of the event
stream, but it restricts parallel execution. Large temporal mini-batches are
attractive for improving computational throughput, and scalable TGN systems
have consequently developed increasingly optimized batching, sampling,
memory-management, and distributed-training mechanisms
\cite{zhou2022tgl,zhou2023disttgl}.


Batch-parallel processing introduces a particular difficulty for memory-based
TGNs when the same node occurs multiple times within a batch. Consider a node
\(u\) that participates in several interactions at different positions of the
same temporal batch. When these interactions are processed in parallel, a
later occurrence may be evaluated before the state changes induced by its
earlier same-batch interactions have been incorporated into persistent
memory. Consequently, multiple interactions can rely on the same pre-batch
memory even though they occur at different points in the temporal sequence.
This loss of intra-batch temporal dependency has been identified as
\emph{temporal discontinuity} and becomes increasingly consequential as
temporal batch size grows \cite{su2024pres}. Recent work further shows that
intra-batch information loss and outdated memory states can substantially
affect the accuracy of memory-based TGNs under batch-parallel training
\cite{feng2025rbt}.


We formulate this issue as a problem of \emph{missing intra-batch state
evolution}. At the beginning of a temporal batch, the last committed memory
of node \(u\) provides a reliable representation of its history up to the
batch boundary. We refer to this state as the node's \emph{pre-batch anchor}.
For a later occurrence of \(u\) within the same batch, however, the memory
required to represent the node at that event should additionally reflect its
causally preceding same-batch interactions. The difference between this
event-specific state and the available pre-batch anchor therefore corresponds
to state evolution that has occurred in the temporal sequence but has not yet
been materialized by the persistent recurrent memory.


Several recent approaches attempt to alleviate this tension between memory
dependency and scalable execution. TGL and DistTGL improve the systems
infrastructure for large-scale and distributed temporal graph training
\cite{zhou2022tgl,zhou2023disttgl}. PRES explicitly addresses temporal
discontinuity through an iterative prediction--correction procedure together
with memory-coherence learning \cite{su2024pres}. RBT measures intra-batch
information loss more finely and combines batch refinement with memory
compensation \cite{feng2025rbt}, while Cascade uses dependency-aware
information to adapt temporal batch boundaries and determine how many
revisits can be tolerated before memory inconsistency becomes harmful
\cite{dai2025cascade}. These approaches motivate a complementary question:
\emph{instead of primarily restricting, partitioning, or iteratively
correcting intra-batch dependencies, can the missing state required at each
interaction be reconstructed directly from the causal information already
available within the batch?}


To address this question, we propose \textbf{AnchorRM}, an anchor-based
residual memory mechanism for event-specific intra-batch state
reconstruction. AnchorRM treats the last committed node memory at the
beginning of each batch as a pre-batch anchor. For an interaction at position
\(k\), it uses the node's causally preceding same-batch interactions to infer
the missing change in memory since that anchor. This missing evolution is
represented by a learned residual memory, which calibrates the pre-batch
anchor to reconstruct the node state required at the current interaction.
Thus, rather than predicting an event-specific state from scratch, AnchorRM
estimates only the state change that has not yet been captured by persistent
memory.


A key property of AnchorRM is that these event-specific states are not
reconstructed recursively. For event \(k\), the residual is inferred directly
from the fixed pre-batch anchor and interactions occurring before \(k\) in
the batch; it does not depend on the reconstructed state of event \(k-1\).
Consequently, once causal same-batch histories have been identified, the
residual corrections for different event positions can be evaluated in
parallel. AnchorRM therefore preserves the causal information boundary of
each interaction while removing the recurrent computational dependency among
the reconstructed intra-batch states.


The resulting architecture operates at two complementary temporal scales.
Persistent node memory captures committed evolution across temporal batches,
whereas residual memory reconstructs the event-specific evolution that is
temporarily missing inside the current batch. The reconstructed endpoint
states are used for interaction prediction and are also incorporated into the
messages that subsequently update persistent memory. AnchorRM therefore
augments, rather than replaces, the underlying recurrent TGN memory
mechanism.


Experiments across four temporal graph datasets and three
memory-based backbones demonstrate the effectiveness of
AnchorRM in large-batch temporal graph learning.
under a relative validation-AP tolerance of $\epsilon=0.01$, AnchorRM
achieves an average $28.67\times$ temporal batch scaling
(up to $40\times$) and $6.16\times$ training speedup
(up to $10.05\times$) relative to the original backbones.
It maintains competitive predictive performance,
achieving an average test AP of 87.43\%, compared with
86.84\% for the original backbones and 86.97\% for PRES.
These findings suggest that causal intra-batch state
reconstruction enables more aggressive temporal batching
while mitigating predictive degradation.


Our main contributions are as follows:
\begin{itemize}

\item We formulate the stale-memory problem in batch-parallel
memory-based TGNs as an \emph{event-specific missing-state problem},
where the unavailable intra-batch evolution is represented relative to
the last committed pre-batch memory.

\item We introduce \textbf{AnchorRM}, an anchor-based residual memory
mechanism that uses causally preceding same-batch interactions to infer
the missing state change and reconstruct the node memory required at
each interaction.

\item We develop a causal parallel formulation in which each
event-specific residual is inferred directly from a shared pre-batch
anchor and its corresponding causal history, eliminating recurrent
dependencies among reconstructed states within a temporal batch.

\item We conduct extensive experiments across four
datasets and three memory-based backbones,
demonstrating substantial improvements in
accuracy-tolerant batch scaling and training
efficiency. Ablation studies further investigate
the effects of causal history capacity,
residual reconstruction, and model dimensions.

\end{itemize}

\section{Related Work}
\label{sec:related}


\subsection{Memory-Based Temporal Graph Neural Networks}
%
Continuous-time temporal graph models represent dynamic systems as streams of
timestamped interactions and update node representations as new events arrive.
JODIE models user and item trajectories using coupled recurrent neural networks
and introduces t-Batch to construct time-consistent batches for efficient
training \citep{kumar2019jodie}. DyRep models evolving node representations
while jointly capturing communication and structural dynamics in dynamic
graphs \citep{trivedi2019dyrep}. Temporal Graph Networks (TGN) provide a more
general framework that combines node memory, message construction, temporal
encoding, message aggregation, memory updating, and graph-based embedding
modules \citep{rossi2020tgn}.
Persistent node memories in these models are closely related to recurrent
state modeling. Gated recurrent units (GRUs) update a hidden state by combining
the previous state with newly observed information through learned gating
mechanisms \citep{cho2014gru}. In memory-based temporal graph models, recurrent
updaters such as GRUs can therefore be used to evolve persistent node states
from temporally aggregated interaction messages.

\subsection{Mitigating Intra-Batch Temporal Staleness} 
%
Several systems have been developed to improve the scalability of temporal
graph neural networks. TGL provides a general framework for large-scale TGN
training with optimized temporal sampling, mailbox storage, node-memory
management, message passing, and scheduling mechanisms
\citep{zhou2022tgl}. Its random chunk scheduling strategy is designed to
reduce the impact of obsolete memories when large temporal batches are used.
DistTGL extends memory-based TGN training to distributed GPU environments and
addresses synchronization and communication challenges associated with
distributed node memories \citep{zhou2023disttgl}.

Recent work has also studied the loss of temporal information introduced by
batch-parallel processing. PRES characterizes the neglect of dependencies
among temporally related interactions processed within the same batch as
\emph{temporal discontinuity} and introduces an iterative
prediction--correction mechanism together with memory-coherence learning
\citep{su2024pres}. RBT formulates the problem in terms of intra-batch
information loss, develops a finer-grained measure of this loss, and combines
refined batch construction with memory compensation
\citep{feng2025rbt}. Cascade adopts a dependency-aware batching strategy that
uses spatial independence and node-memory stability to determine when temporal
dependencies can be relaxed while maintaining efficient execution
\citep{dai2025cascade}.
\subsection{Aggregation in Graph Neural Networks}

Aggregation is a central operation in graph neural networks because a node
representation is commonly constructed by combining information from multiple
local signals. GraphSAGE develops inductive representation learning through
sampling and learned neighborhood aggregation
\citep{hamilton2017graphsage}. Graph Attention Networks (GAT) introduce
attention-based aggregation, allowing neighboring nodes to contribute with
different learned importance weights \citep{velickovic2018gat}. Graph
Isomorphism Networks (GIN) study the expressive power of neighborhood
aggregation and show the importance of expressive aggregation functions for
distinguishing different graph structures \citep{xu2019gin}.

These works illustrate several major approaches to combining multiple inputs,
including learned neighborhood aggregation, attention-based weighting, and
expressive multiset aggregation. Such aggregation mechanisms are particularly
relevant when models must combine a variable number of interaction or
neighborhood signals into a fixed-dimensional representation.

\subsection{Residual Memory Modeling} 
%
%
Residual modeling has also been explored in sequential learning. VARNN
introduces an explicit residual-memory pathway in which recent prediction
deviations are represented separately from the primary predictive state and
used to recalibrate subsequent predictions \citep{gharwi2025varnn}. This
primary-state/residual-state decomposition provides a mechanism for separating
the dominant temporal representation from additional information associated
with recent deviations or variability.
AnchorRM follows the broader principle of maintaining complementary primary
and residual state information, but assigns the residual a different semantic
role. Rather than representing prediction error, the residual in AnchorRM
models the node-state evolution that is missing from the persistent memory due
to causally preceding interactions within the current temporal batch. The
persistent node memory therefore serves as the primary state, while the
residual captures the uncommitted intra-batch evolution required to reconstruct
an event-specific node state.

\section{Preliminaries and Problem Formulation}
\label{sec:problem}

\subsection{Temporal Interaction Graphs}

We consider a continuous-time temporal graph represented by a chronologically
ordered sequence of interactions
$
\mathcal{E}
=
\{e_1,e_2,\ldots,e_T\},
$
where each interaction is defined as
$
e_k=(u_k,v_k,t_k,x_k).
$
Here, \(u_k\) and \(v_k\) denote the source and destination nodes involved in
the \(k\)-th interaction, \(t_k\) denotes its timestamp, and
\(x_k\in\mathbb{R}^{d_x}\) represents optional interaction features. We assume
the interaction stream is processed in chronological order,
$
t_1\leq t_2\leq\cdots\leq t_T.
$
The learning task considered in this work is temporal link prediction, where
the model predicts whether an interaction between a pair of nodes occurs at a
given point in the event stream using only information available before that
interaction.

\subsection{Persistent Memory in Temporal Graph Networks}
Memory-based temporal graph neural networks maintain a persistent state for
each node to summarize its historical interactions
\citep{rossi2020tgn}. Let
$
h_u\in\mathbb{R}^{d_h}
$
denote the persistent memory associated with node \(u\). When a new
interaction involving \(u\) is processed, a message \(m_u\) is constructed
from the current interaction and relevant node states. An aggregated message
is then used to update the node memory through a recurrent memory updater,
$
h_u^{\mathrm{new}}
=
U(h_u^{\mathrm{old}},m_u),
$
where \(U(\cdot)\) may be implemented using a gated recurrent unit (GRU)
\citep{cho2014gru}.

This recurrent formulation naturally induces a temporal dependency between
successive interactions involving the same node. 

\subsection{Temporal Mini-Batching}

For computational efficiency, temporal interactions are commonly processed in
chronological mini-batches. Let
$
\mathcal{B}
=
\{e_1,e_2,\ldots,e_B\}
$
denote a temporal batch containing \(B\) consecutive interactions. For a node
\(u\), let
$
a_u
=
h_u^{\mathrm{pre}}
$
denote its last committed persistent memory immediately before processing the
batch. We refer to \(a_u\) as the \emph{pre-batch state}.
If node \(u\) occurs only once in the batch, this pre-batch state contains all
of its historical information available before that interaction, except for
events involving other nodes that do not affect \(u\)'s recurrent state. The
difficulty arises when \(u\) occurs multiple times within the same batch.

Suppose node \(u\) participates in interactions at batch positions
$
j_1<j_2<\cdots<j_r<k.
$
Under exact sequential recurrent processing, the state required immediately
before event \(e_k\) would be

$$
h_{u,k}^{*}
=
U\Big(
\cdots
U(
U(a_u,m_{u,j_1}),
m_{u,j_2}
)
\cdots,
m_{u,j_r}
\Big),
$$

where \(m_{u,j_i}\) denotes the message associated with node \(u\) at
interaction position \(j_i\).
Therefore,
$
h_{u,k}^{*}
$
contains the accumulated effect of the relevant same-batch interactions that
occurred before \(e_k\).

\subsection{Intra-Batch Stale Memory}

Batch-parallel processing can prevent these intermediate recurrent states from
being materialized before later interactions in the same batch are evaluated.
In this case, multiple occurrences of node \(u\) can use a state that remains
close to the pre-batch memory,
$
h_{u,k}^{\mathrm{batch}}
\approx
a_u,
$
even though the sequentially correct state is \(h_{u,k}^{*}\).
Consider three interactions involving node $A$:

$$
e_1=(A,B),\qquad
e_2=(A,C),\qquad
e_3=(A,D),
$$

the exact temporal dependency is

$$
h_A^{(0)}
\xrightarrow{e_1}
h_A^{(1)}
\xrightarrow{e_2}
h_A^{(2)}
\xrightarrow{e_3}
h_A^{(3)}.
$$

However, under large-batch processing, prediction at the three event positions
may effectively use

$$
e_1:\;h_A^{(0)},
\qquad
e_2:\;h_A^{(0)},
\qquad
e_3:\;h_A^{(0)},
$$

before the corresponding same-batch updates are committed.
For \(e_3\), the discrepancy is especially clear: the state required for
prediction should contain the effects of \(e_1\) and \(e_2\), but the
available persistent memory may contain neither. This phenomenon is closely
related to the temporal discontinuity and intra-batch information-loss
problems identified in prior work \citep{su2024pres,feng2025rbt}.

\subsection{Missing Intra-Batch State Evolution}

Rather than viewing stale memory only as an outdated-state problem, we
formulate it in terms of the state evolution that is missing from the
pre-batch memory.
For node \(u\) at event position \(k\), define the ideal missing state
evolution as

$$
r_{u,k}^{*}
=
h_{u,k}^{*}-a_u.
$$

The quantity \(r_{u,k}^{*}\) represents the cumulative effect of the causally
preceding same-batch interactions involving node \(u\) that has not yet been
reflected in its persistent memory.
For the first occurrence of \(u\) in a batch, no earlier same-batch
interaction involving \(u\) exists, and therefore
$
r_{u,k}^{*}=0.
$
For later occurrences, \(r_{u,k}^{*}\) generally becomes nonzero and depends
on the sequence of preceding interactions involving \(u\).
Let the causal same-batch history available to node \(u\) before event
position \(k\) be

$$
\mathcal{H}_{u,k}
=
\left\{
q_{u,j}
\;:\;
j<k,\;
u\in e_j
\right\},
$$

where \(q_{u,j}\) denotes a representation of interaction \(e_j\) from the
perspective of node \(u\). By construction, \(\mathcal{H}_{u,k}\) contains
only interactions that occur earlier than \(e_k\) in the temporal batch and
therefore preserves event-order causality.

\subsection{Problem Formulation}

The objective is to recover the event-specific node state required at position
\(k\) without explicitly executing the corresponding sequence of recurrent
memory updates inside the batch.
Equivalently, given the pre-batch state \(a_u\) and the causal same-batch
history \(\mathcal{H}_{u,k}\), we seek a function

$$
R_{\theta}
\left(
a_u,
\mathcal{H}_{u,k}
\right)
$$

that approximates the missing state evolution,
$
R_{\theta}
\left(
a_u,
\mathcal{H}_{u,k}
\right)
\approx
r_{u,k}^{*}
=
h_{u,k}^{*}-a_u.
$
Then, the corresponding reconstructed event-specific state is:

$$
\tilde h_{u,k}
=
a_u+
R_{\theta}
\left(
a_u,
\mathcal{H}_{u,k}
\right).
$$

Thus, she desired reconstruction should satisfy three properties:

\begin{enumerate}

\item \textbf{Causality.}
The reconstructed state for event \(k\) must depend only on information
available before \(e_k\). In particular, neither the current event nor
future events may enter \(\mathcal{H}_{u,k}\).

\item \textbf{Event specificity.}
Repeated occurrences of the same node within a batch should be allowed to
receive different reconstructed states according to their respective
causal histories.

\item \textbf{Parallelizability.}
Reconstruction of the state at position \(k\) should not require the
previously reconstructed state at position \(k-1\). Removing this
recurrent dependency permits event-specific state reconstruction to be
evaluated in parallel after the causal histories have been constructed.

\end{enumerate}

This formulation transforms large-batch stale memory into a causal residual
state-reconstruction problem. The next section introduces the proposed
AnchorRM mechanism for estimating this missing intra-batch evolution.

\section{AnchorRM: Anchor-Based Residual Memory Reconstruction}
\label{sec:method}

\subsection{Overview}

The formulation in Section~\ref{sec:problem} identifies the unavailable
intra-batch state evolution for node \(u\) at event position \(k\) as
$
r_{u,k}^{*}
=
h_{u,k}^{*}-a_u,
$
where \(a_u\) is the node state committed before the current temporal batch and
\(h_{u,k}^{*}\) is the state that would ideally incorporate the effects of all
causally preceding same-batch interactions involving \(u\).
We propose \textbf{AnchorRM}, an anchor-based residual memory mechanism that
directly estimates this missing evolution without executing the corresponding
recurrent updates sequentially. For each node, the last committed persistent
memory at the beginning of the batch is treated as a fixed \emph{anchor}
within that batch. Causally preceding same-batch interactions are encoded into
event tokens and used to estimate an event-specific residual,

$$
\hat{\mathbf r}_{u,k}
=
R_{\theta}
\left(
a_u,\mathcal H_{u,k}
\right).
$$

The node state used at event \(k\) is reconstructed as

$$
\tilde h_{u,k}
=
a_u+\hat{\mathbf r}_{u,k}
.
$$

The reconstructed state therefore consists of two components: the committed
historical information contained in \(a_u\), and an estimated correction for
the state evolution that has occurred after that anchor but has not yet been
committed to persistent memory.

\subsection{Pre-Batch Memory Anchor}

Consider a temporal batch
$
\mathcal B
=
\{e_1,e_2,\ldots,e_B\}.
$
Before processing the interactions in \(\mathcal B\), AnchorRM retrieves the
persistent memory of every node required by the batch. For node \(u\), we
define
$
a_u
=
h_u^{\mathrm{pre}}
$
as its \emph{pre-batch anchor}. We additionally retain the node's most recent
committed update time, denoted by
$
\tau_u.
$
The anchor remains unchanged while event-specific states for the current batch
are reconstructed. Importantly, it is fixed only within the current batch.
After the batch has been processed and its messages have been committed, the
updated persistent memory becomes the anchor for the next temporal batch.

Using a shared pre-batch anchor provides a stable reference point from which
the missing state evolution at different event positions can be estimated
independently.

\subsection{Anchor-Conditioned Interaction Tokens}

The residual required at a later interaction depends not only on how many
earlier interactions involved the node, but also on the content and temporal
context of those interactions. We therefore encode each interaction into a
compact node-oriented event token.
For an interaction

$$
e_k=(u_k,v_k,t_k,x_k),
$$

the source-oriented token is constructed as
$
q_{u_k,k}
=
E_{\phi}
\left(
\left[
a_{u_k},
a_{v_k},\allowbreak
x_k,
\operatorname{TE}(t_k-\tau_{u_k})
\right]
\right),
$
where \(E_{\phi}\) is a learned event encoder and
\(\operatorname{TE}(\cdot)\) denotes a temporal encoding.
Similarly, the destination-oriented token reverses the endpoint roles,
$
q_{v_k,k}
=
E_{\phi}
\left(
\left[
a_{v_k},
a_{u_k},
x_k,
\operatorname{TE}(t_k-\tau_{v_k})
\right]
\right).
$
Thus, the same interaction produces different representations from the
perspectives of its two participating nodes. The node's own anchor provides
its historical context, while the partner anchor describes the state of the
node with which it interacted.
All event tokens in the batch are constructed from the pre-batch anchor
snapshot. Their construction therefore does not require sequential
intra-batch memory updates.

\subsection{Causal Same-Batch Interaction History}

For node \(u\) at event position \(k\), AnchorRM considers only interactions
involving \(u\) that occur earlier in the same batch:

$$
\mathcal H_{u,k}
=
\left[
q_{u,j_1},
q_{u,j_2},
\ldots,
q_{u,j_r}
\right],
$$

where
$
j_1<j_2<\cdots<j_r<k.
$
The strict condition
$
j<k
$
is central to the construction. The token associated with the current event
\(e_k\) is not included in the history used to reconstruct the state for
\(e_k\), and neither are tokens from future event positions.
For the example introduced in Section~\ref{sec:problem},
the causal histories of node $A$ are

$$
\mathcal H_{A,1}=\varnothing, \qquad
\mathcal H_{A,2}=[q_{A,1}], \qquad
\mathcal H_{A,3}=[q_{A,1},q_{A,2}].
$$

Consequently, the reconstructed state used for \(e_3\) can reflect the effects
of \(e_1\) and \(e_2\), but it cannot access information from \(e_3\) itself.

\subsection{Fixed-Capacity Causal History with Masking}

Different nodes can have different numbers of preceding interactions within a
batch. To support tensorized computation, AnchorRM converts each
variable-length causal history into a fixed-size representation.
Let \(K\) denote the maximum number of preceding interaction tokens retained
for each query. If more than \(K\) causal events are available, the most recent
\(K\) are retained while preserving their chronological order. The resulting
history tensor is
$
H_{u,k}
\in
\mathbb{R}^{K\times d_q},
$

where \(d_q\) is the event-token dimension.
A corresponding binary mask
$
M_{u,k}
\in
\{0,1\}^{K}
$

indicates which positions contain valid causal events. For example, if only
three previous interactions are available and \(K=5\),

$$
H_{u,k}
=
\begin{bmatrix}
q_{u,j_1}\\
q_{u,j_2}\\
q_{u,j_3}\\
0\\
0
\end{bmatrix},
\qquad
M_{u,k}
=
[1,1,1,0,0].
$$

The mask serves two purposes. First, it distinguishes valid interaction tokens
from padded positions. Second, together with the history-construction rule
\(j<k\), it ensures that only causally available same-batch information is
presented to the residual estimator.
Masking itself does not determine how the preceding interactions should be
combined. Instead, it specifies which history positions are valid, while the
learned residual estimator determines how the corresponding interaction
signals contribute to the reconstructed state.

\subsection{Residual Memory Estimation}

The masked causal history is combined with the pre-batch anchor to estimate the
missing state evolution. Let

$$
\bar H_{u,k}
=
\operatorname{vec}
\left(
H_{u,k}\odot M_{u,k}
\right)
$$

denote the flattened masked interaction history. AnchorRM computes

$$
\hat{\mathbf r}_{u,k}
=
R_{\theta}
\left(
\left[
a_u,
\bar H_{u,k},
M_{u,k}
\right]
\right),
$$

where \(R_{\theta}\) is a feed-forward residual estimator.

The inclusion of the anchor allows the same interaction history to imply
different corrections depending on the node's committed state. The event
tokens preserve information about the preceding interactions, while the mask
provides the residual estimator with the valid history structure.
If no causally preceding same-batch interaction involving \(u\) exists, no
intra-batch correction is required and we set
$
\hat{\mathbf r}_{u,k}=0.
$
Otherwise, the residual estimator approximates
$
\hat{\mathbf r}_{u,k}
\approx
r_{u,k}^{*}
=
h_{u,k}^{*}-a_u.
$
The event-specific state is consequently reconstructed as
$
\tilde h_{u,k}
=
a_u+\hat{\mathbf r}_{u,k}
.
$

\subsection{Interaction Prediction}

For interaction \(e_k=(u_k,v_k,t_k,x_k)\), AnchorRM reconstructs the
event-specific states of both endpoints,
$
\tilde h_{u_k,k}
=
a_{u_k}+\hat{\mathbf r}_{u_k,k}
$, 
$
\tilde h_{v_k,k}
=
a_{v_k}+\hat{\mathbf r}_{v_k,k}.
$

These states replace the stale pre-batch endpoint memories when computing the
representation used for temporal link prediction. A scoring function
\(S_{\psi}\) produces the interaction score,

$$
\hat y_k
=
S_{\psi}
\left(
\tilde h_{u_k,k},
\tilde h_{v_k,k}
\right).
$$

Negative node pairs are evaluated analogously using the causal history
available to the queried nodes at the corresponding event position.

Therefore, two occurrences of the same node within one temporal batch need
not use the same representation: each occurrence receives an event-specific
state determined by the causal interaction history available at that position.

\subsection{Persistent Memory Update}

The reconstructed states are temporary event-specific representations; they do
not replace the persistent TGN memory. After prediction, they are also used
to construct the messages that will subsequently update persistent memory.
For interaction \(e_k=(u_k,v_k,t_k,x_k)\), the endpoint messages are formed
using the reconstructed states,
$
m_{u_k,k}
=
\left[
\tilde h_{u_k,k},
\tilde h_{v_k,k},
x_k,
\operatorname{TE}(t_k-\tau_{u_k})
\right],
$
and
$
m_{v_k,k}
=
\left[
\tilde h_{v_k,k},
\tilde h_{u_k,k},
x_k,
\operatorname{TE}(t_k-\tau_{v_k})
\right].
$
The resulting messages are aggregated according to the underlying memory-based
TGN and used by its recurrent updater to advance persistent node memory at
the batch boundary.

\subsection{Causality and Parallelization}

AnchorRM separates temporal causality from recurrent computational dependency.
Causality is enforced through the history condition
$
j<k,
$
so the state used for event \(k\) can incorporate earlier same-batch
interactions but cannot incorporate the current or future interactions.

At the same time, every residual has the form
$
\hat{\mathbf r}_{u,k}
=
R_{\theta}
\left(
a_u,
\mathcal H_{u,k}
\right),
$
rather than
$
\hat{\mathbf r}_{u,k}
=
R_{\theta}
\left(
\tilde h_{u,k-1},
q_{u,k-1}
\right).
$
Consequently, after the causal histories have been assembled, residual
estimation across event positions does not require sequential recurrent
execution. The method therefore preserves event-order causality while
allowing intra-batch state reconstruction to be implemented with parallel
tensor operations.

\section{Experimental Setup}
\label{sec:experimental_setup}

We evaluate AnchorRM for temporal link prediction with two objectives.
First, we study whether causal intra-batch state reconstruction reduces
the predictive degradation caused by increasingly large temporal
batches. Second, we want to examine how much larger a temporal batch AnchorRM can support to the best-performing TGN configuration, and measure the corresponding training speedup. Unless otherwise stated, all main experiments use the default
AnchorRM configuration $K=4$, event-token dimension $d_q=16$, and
residual hidden dimension $d_r=64$. Variations of these architectural
parameters are reserved for the ablation study.

\subsection{Datasets}

We evaluate AnchorRM on four temporal interaction benchmarks:
Wikipedia, Reddit, MOOC, and LastFM \citep{kumar2019jodie}. 
We follow the chronological preprocessing and inductive evaluation protocol used by TGN, preserving
the temporal ordering of all interactions.

\subsection{Baselines}
We compare AnchorRM with established memory-based temporal
graph learning methods and existing approaches for mitigating
intra-batch temporal dependencies. Our primary baseline is
Temporal Graph Networks (TGN) \citep{rossi2020tgn}, which
maintains persistent node memories through recurrent updates.
To investigate generalization across memory-based architectures,
we additionally consider JODIE \citep{kumar2019jodie},
DyRep \citep{trivedi2019dyrep}.
We further evaluate the PRES framework \citep{su2024pres},
which addresses temporal discontinuity through iterative
prediction--correction and memory-coherence learning, using
its official repository and the default experimental settings
reported in the original paper.
For controlled comparisons, we use consistent data splits,
negative sampling, training budgets, and checkpoint-selection
criteria. TGN, JODIE, DyRep, and AnchorRM are evaluated using
the same tensorized execution backend.

\subsection{Training Protocol}

We train the evaluated backbones and their AnchorRM
variants using a learning rate of $10^{-4}$ and
consistent chronological data splits. Each experiment was conducted on a single NVIDIA Tesla V100 GPU. 
Unless otherwise
specified, the main batch-scaling experiments use
a maximum of 100 training epochs, with checkpoints
selected according to validation AP.
We evaluate temporal batch sizes
$B\in\{100,200,500,1000,2000,4000\}$.
For each backbone, the training configuration is
kept fixed across batch sizes to isolate the effect
of temporal batching.

Since persistent node memory represents part of
the model's temporal state, checkpoints preserve
both the learned parameters and the corresponding
memory state. Evaluation follows chronological
order, with testing initialized from the memory
state obtained after processing validation.
Overall and new-node evaluations use separate
memory backups to prevent interference between
evaluation settings.

All main experiments use random seed 2026,
with consistent data splits, fixed evaluation
negative samplers, and matched optimization
settings across the original backbones and
their AnchorRM variants.
\begin{figure*}[!t]
\centering
\includegraphics[width=0.80\textwidth]
{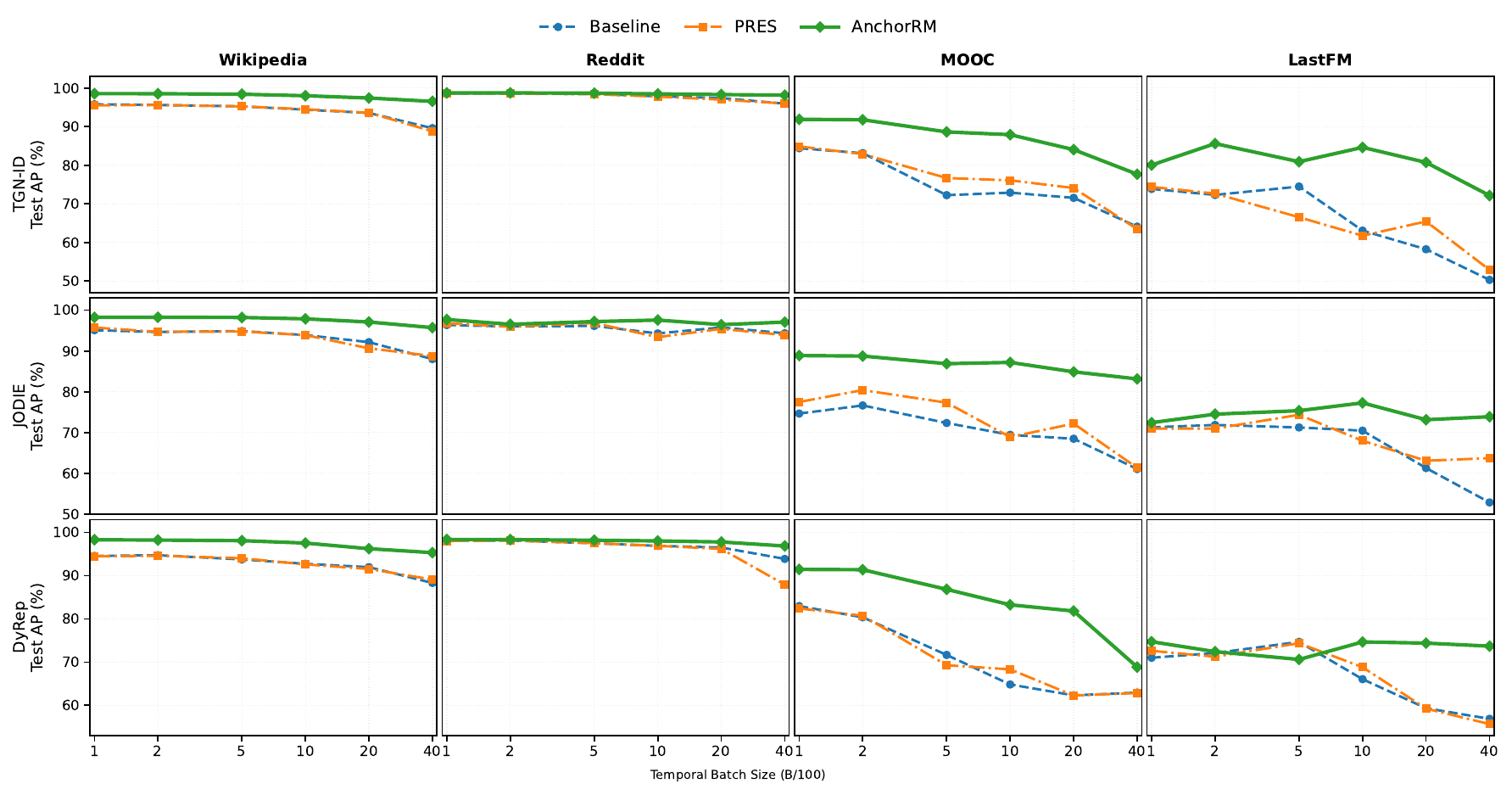}
\caption{
Test AP across temporal batch sizes on Wikipedia, Reddit,
MOOC, and LastFM using TGN-ID, JODIE, and DyRep backbones.
AnchorRM generally maintains higher predictive performance
as temporal batch sizes increase compared with the original
backbones and PRES.
}
\label{fig:batch_accuracy}
\end{figure*}

\newcommand{\BatchCell}[2]{%
  \mbox{$#1\times$\,{\scriptsize (#2)}}%
}

\newcommand{\SpeedCell}[2]{%
  \mbox{$#1\times$\,{\scriptsize (#2s)}}%
}

\begin{table*}[!t]
\centering

\caption{
Accuracy-tolerant batch scaling and training efficiency
($\epsilon=0.01$). Bold indicates the highest test AP
within each backbone and dataset.
Numbers in parentheses indicate batch size or seconds per epoch.
}

\label{tab:main_results}

\renewcommand{\arraystretch}{1.12}
\setlength{\tabcolsep}{3pt}

\resizebox{\textwidth}{!}{%
\begin{tabular}{@{}l|ccc|ccc|ccc|ccc@{}}

\toprule

& \multicolumn{3}{c|}{\textbf{REDDIT}}
& \multicolumn{3}{c|}{\textbf{WIKI}}
& \multicolumn{3}{c|}{\textbf{MOOC}}
& \multicolumn{3}{c}{\textbf{LASTFM}} \\

\cmidrule(lr){2-4}
\cmidrule(lr){5-7}
\cmidrule(lr){8-10}
\cmidrule(l){11-13}

\textbf{Model}
& \textbf{AP(\%)}
& \textbf{Batch}
& \textbf{Speedup}
& \textbf{AP(\%)}
& \textbf{Batch}
& \textbf{Speedup}
& \textbf{AP(\%)}
& \textbf{Batch}
& \textbf{Speedup}
& \textbf{AP(\%)}
& \textbf{Batch}
& \textbf{Speedup} \\

\midrule


TGN-ID
& \textbf{98.62}
& \BatchCell{1}{200}
& \SpeedCell{1.00}{8.84}
& 95.81
& \BatchCell{1}{100}
& \SpeedCell{1.00}{2.45}
& \textbf{84.42}
& \BatchCell{1}{100}
& \SpeedCell{1.00}{7.61}
& 73.87
& \BatchCell{1}{100}
& \SpeedCell{1.00}{18.09} \\

TGN-ID+PRES$^{\dagger}$
& 97.79
& \BatchCell{5}{1000}
& \SpeedCell{3.91}{2.26}
& 95.25
& \BatchCell{5}{500}
& \SpeedCell{3.60}{0.68}
& 82.89
& \BatchCell{2}{200}
& \SpeedCell{1.71}{4.46}
& \textbf{74.38}
& \BatchCell{1}{100}
& \SpeedCell{0.91}{19.98} \\

TGN-ID+AnchorRM
& 98.15
& \BatchCell{20}{4000}
& \SpeedCell{5.53}{1.60}
& \textbf{96.55}
& \BatchCell{40}{4000}
& \SpeedCell{9.61}{0.25}
& 84.04
& \BatchCell{20}{2000}
& \SpeedCell{10.05}{0.76}
& 72.15
& \BatchCell{40}{4000}
& \SpeedCell{9.00}{2.01} \\

\midrule


JODIE
& 96.00
& \BatchCell{1}{200}
& \SpeedCell{1.00}{6.96}
& 95.10
& \BatchCell{1}{100}
& \SpeedCell{1.00}{2.31}
& 76.64
& \BatchCell{1}{200}
& \SpeedCell{1.00}{3.32}
& 71.28
& \BatchCell{1}{100}
& \SpeedCell{1.00}{17.41} \\

JODIE+PRES$^{\dagger}$
& 95.42
& \BatchCell{10}{2000}
& \SpeedCell{6.73}{1.03}
& 94.77
& \BatchCell{5}{500}
& \SpeedCell{3.81}{0.61}
& 80.41
& \BatchCell{1}{200}
& \SpeedCell{0.84}{3.96}
& \textbf{74.35}
& \BatchCell{5}{500}
& \SpeedCell{4.23}{4.12} \\

JODIE+AnchorRM
& \textbf{97.08}
& \BatchCell{20}{4000}
& \SpeedCell{4.56}{1.53}
& \textbf{95.73}
& \BatchCell{40}{4000}
& \SpeedCell{9.26}{0.25}
& \textbf{83.15}
& \BatchCell{20}{4000}
& \SpeedCell{5.77}{0.58}
& 73.87
& \BatchCell{40}{4000}
& \SpeedCell{8.42}{2.07} \\

\midrule


DyRep
& \textbf{98.12}
& \BatchCell{1}{100}
& \SpeedCell{1.00}{19.53}
& 94.55
& \BatchCell{1}{100}
& \SpeedCell{1.00}{3.52}
& \textbf{82.95}
& \BatchCell{1}{100}
& \SpeedCell{1.00}{9.72}
& \textbf{74.68}
& \BatchCell{1}{500}
& \SpeedCell{1.00}{10.66} \\

DyRep+PRES$^{\dagger}$
& 97.51
& \BatchCell{5}{500}
& \SpeedCell{2.50}{7.81}
& 94.03
& \BatchCell{5}{500}
& \SpeedCell{2.42}{1.46}
& 82.41
& \BatchCell{1}{100}
& \SpeedCell{0.88}{11.02}
& 74.37
& \BatchCell{1}{500}
& \SpeedCell{0.96}{11.12} \\

DyRep+AnchorRM
& 96.86
& \BatchCell{40}{4000}
& \SpeedCell{3.60}{5.43}
& \textbf{95.32}
& \BatchCell{40}{4000}
& \SpeedCell{3.40}{1.04}
& 81.81
& \BatchCell{20}{2000}
& \SpeedCell{3.56}{2.73}
& 74.39
& \BatchCell{4}{2000}
& \SpeedCell{1.17}{9.11} \\

\bottomrule

\end{tabular}%
}

\vspace{3pt}

\parbox{\textwidth}{\footnotesize
}

\end{table*}

\subsection{Evaluation Metrics}

We report average precision (AP) for both overall
temporal link prediction and inductive new-node
prediction, where new nodes are those not observed
during training \citep{rossi2020tgn}.
AP is used for checkpoint selection and predictive
performance evaluation. Fixed evaluation negative
samplers are used across compared methods to
ensure consistent evaluation.

\subsubsection{Accuracy-Tolerant Batch Scaling and Training Speedup}
\label{sec:batch_scaling}
Increasing temporal batch size can improve training
efficiency but may degrade predictive performance
due to intra-batch temporal dependencies.
To quantify this accuracy–efficiency trade-off, we evaluate how much the temporal batch size can be increased while
maintaining a prescribed level of predictive quality. We define the
\emph{accuracy-tolerant batch-scaling factor} as follows.
Let
$
\mathcal{B}=\{100,200,500,1000,2000,4000\}
$
denote the set of evaluated temporal batch sizes. For each backbone and
dataset, we first select the reference configuration according to
validation average precision (AP):

\begin{equation}
B_{\mathrm{base}}^{*}
=
\arg\max_{B\in\mathcal{B}}
\mathrm{AP}^{\mathrm{val}}_{\mathrm{base}}(B).
\end{equation}

The corresponding reference validation accuracy is

\begin{equation}
A_{\mathrm{base}}^{*}
=
\mathrm{AP}^{\mathrm{val}}_{\mathrm{base}}
\left(B_{\mathrm{base}}^{*}\right).
\end{equation}

Let $\epsilon\geq 0$ denote the maximum allowable relative
degradation in validation AP. For each method
$M\in\{\mathrm{PRES},\mathrm{AnchorRM}\}$,
we define the largest admissible batch size as

\begin{equation}
B_M^{*}(\epsilon)
=
\max
\left\{
B\in\mathcal{B}:
\mathrm{AP}^{\mathrm{val}}_M(B)
\geq
(1-\epsilon)A_{\mathrm{base}}^{*}
\right\}.
\end{equation}

The resulting accuracy-tolerant batch-scaling factor is

\begin{equation}
S_B(M,\epsilon)
=
\frac{
B_M^{*}(\epsilon)
}{
B_{\mathrm{base}}^{*}
}.
\end{equation}

We additionally measure the corresponding training speedup as

\begin{equation}
S_T(M,\epsilon)
=
\frac{
T_{\mathrm{base}}(B_{\mathrm{base}}^{*})
}{
T_M(B_M^{*}(\epsilon))
},
\end{equation}

where $T_M(B)$ denotes the mean training time per epoch
at batch size $B$, excluding validation and testing.
We set $\epsilon=0.01$, corresponding to a 1\% relative
validation-AP tolerance. Batch sizes are selected using
validation AP only, while the corresponding test AP is
reported for final evaluation.

\section{Main Results}
\label{sec:results}

We evaluate the predictive performance and
training scalability of AnchorRM across three
memory-based backbones and four temporal graph
datasets. We first examine the effect of temporal
batch size on test AP, followed by accuracy-tolerant
batch scaling, training speedup, and inductive
new-node performance.

\begin{figure*}[!t]
\centering
\includegraphics[width=\textwidth]
{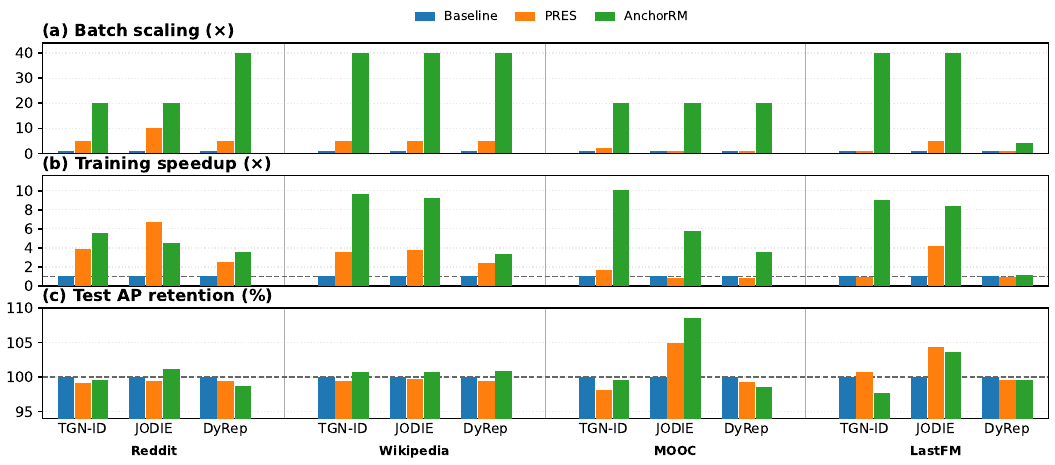}
\caption{
Overall accuracy-tolerant batch scaling and training
efficiency under a relative validation-AP tolerance of $\epsilon=0.01$.
(a) Temporal batch-scaling factor.
(b) Training speedup relative to the original backbone.
(c) Test AP retention relative to the baseline, where
100\% indicates equal predictive performance.
Results cover four datasets and three memory-based
temporal graph backbones.
}
\label{fig:overall_performance}
\end{figure*}

\subsection{Overall Predictive Performance}
Table~\ref{tab:main_results} summarizes the performance of
AnchorRM against the original memory-based temporal graph
backbones and PRES across four datasets under a 1\% relative
validation-AP tolerance. AnchorRM achieves an average
$6.16\times$ training speedup (up to $10.05\times$) and
$28.67\times$ batch scaling (up to $40\times$), while
maintaining competitive predictive performance. Across the
12 backbone--dataset combinations, AnchorRM achieves an
average test AP of 87.43\%, compared with 86.84\% for the
baseline and 86.97\% for PRES.

Figure~\ref{fig:batch_accuracy} illustrates the effect of
increasing temporal batch sizes on test AP across the three
backbones and four datasets. While the original backbones
and PRES generally exhibit substantial performance degradation
at larger batch sizes, AnchorRM maintains more stable
predictive performance, particularly on Wikipedia and MOOC.
This suggests that residual-based memory calibration helps
mitigate the degradation associated with large-batch
temporal graph processing.

Figure~\ref{fig:overall_performance} further highlights
the accuracy--efficiency trade-off, where PRES achieves
an average $3.83\times$ batch scaling and $2.71\times$
training speedup, compared with AnchorRM's substantially
larger gains. AnchorRM achieves higher test AP than the
baseline across all three backbones on Wikipedia and
across all four datasets with JODIE. The largest
improvement occurs on JODIE/MOOC, where test AP increases
from 76.64\% to 83.15\% (+6.51 percentage points),
while increasing the temporal batch size from 200 to
4000 ($20\times$). Overall, these findings demonstrate
that AnchorRM enables substantially more efficient
large-batch training while maintaining competitive
predictive accuracy.

\begin{figure}[t]
\centering
\includegraphics[width=\columnwidth]
{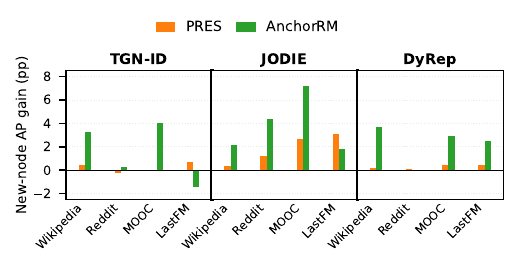}
\caption{
New-node test AP improvement relative to the original
backbones under a relative validation-AP tolerance of $\epsilon=0.01$.
}
\label{fig:new_node_performance}
\end{figure}
\subsection{Performance on New Nodes}

Figure~\ref{fig:new_node_performance} shows that AnchorRM
improves inductive new-node prediction, achieving an
average test AP of 90.36\%, compared with 87.80\%
for the baseline and 88.59\% for PRES.
AnchorRM outperforms the baseline in 10 of 12 configurations,
with the largest improvement on JODIE/MOOC
(+7.22 percentage points) at $20\times$ batch scaling.
These findings suggest that reconstructing event-specific
states from causally preceding intra-batch interactions
provides useful context for nodes with limited
historical memory.


\begin{figure*}[!t]
\centering
\includegraphics[width=0.85\textwidth]
{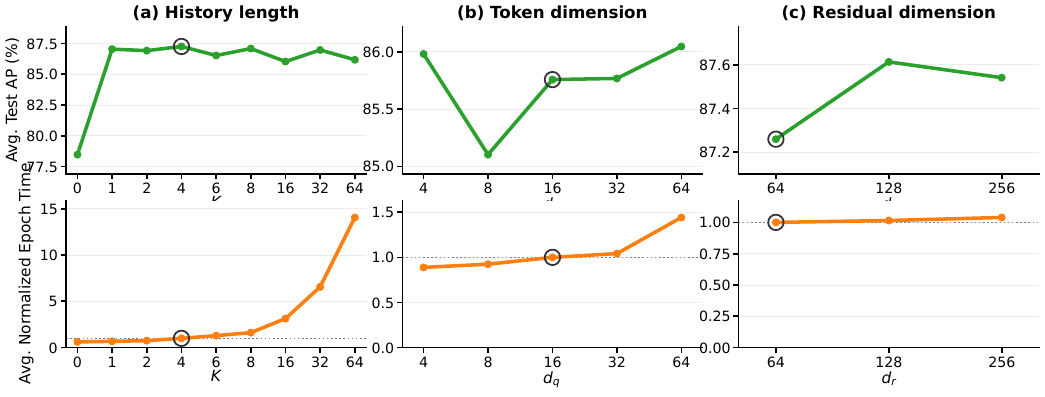}
\caption{
Sensitivity of AnchorRM to (a) causal history length
$K$, (b) event-token dimension $d_q$, and
(c) residual hidden dimension $d_r$.
The top row reports average test AP, while the bottom
row reports normalized training time relative to
the default configuration.
Results are averaged across 12 backbone--dataset
combinations at $B=4000$.
}
\label{fig:capacity_sensitivity}
\end{figure*}

\section{Ablation Studies}
\label{sec:ablation}

\subsection{Architectural Sensitivity}
\label{sec:ablation}

We investigate the sensitivity of AnchorRM to three
architectural parameters: the number of historical
interactions $K$, the event-token dimension $d_q$,
and the residual hidden dimension $d_r$.
Each parameter is varied independently while
the others are fixed at their default values
$(K=4,d_q=16,d_r=64)$, using a temporal batch
size of $B=4000$.
Figure~\ref{fig:capacity_sensitivity} reports
average test AP and normalized epoch time across
12 backbone--dataset combinations.
Epoch time is normalized relative to the default
configuration for each backbone--dataset pair
before averaging.

\paragraph{History Length ($K$).}
Increasing $K$ from 0 to 1 improves average test AP
from 78.47\% to 87.04\%, highlighting the importance
of causal intra-batch history. Performance reaches
87.26\% at $K=4$, while longer histories provide
no consistent improvement and substantially increase
computational overhead. At $K=64$, average test AP
decreases to 86.18\% despite considerably higher
training cost.

\paragraph{Event-Token Dimension ($d_q$).}
Increasing $d_q$ from 16 to 64 yields only a marginal
improvement in average test AP, from 85.76\% to
86.05\%, while increasing epoch time by approximately
31\%. This indicates that larger event representations
provide limited additional predictive benefits
relative to their computational overhead.

\paragraph{Residual Hidden Dimension ($d_r$).}
Increasing $d_r$ from 64 to 128 improves average
test AP from 87.26\% to 87.61\%, with negligible
additional per-epoch cost. Further increasing $d_r$
to 256 slightly reduces test AP to 87.54\%.
These results suggest that increasing residual
reconstruction capacity provides modest benefits,
with diminishing returns at larger dimensions.

Overall, the results indicate that short causal
histories and compact representations provide
a favorable accuracy--efficiency trade-off,
while increasing architectural capacity offers
limited additional gains.

\subsection{Residual vs. Direct State Reconstruction}
\label{sec:ablation_reconstruction}

We investigate whether AnchorRM benefits specifically from reconstructing the intra-batch state as
a residual relative to the fixed pre-batch anchor, rather than predicting the complete state directly.
To isolate this design choice, we keep the anchor, causal same-batch history, reconstruction network,
and training configuration fixed, and vary only the reconstruction rule:

\begin{equation}
\tilde{\mathbf h}_{u,k} =
\begin{cases}
\mathbf a_u + F_{\theta}(\mathbf a_u,\mathcal H_{u,k}),
& \text{Residual}, \\[1mm]
F_{\theta}(\mathbf a_u,\mathcal H_{u,k}),
& \text{Direct}, \\[1mm]
\mathbf a_u +
\alpha\left(
F_{\theta}(\mathbf a_u,\mathcal H_{u,k})-\mathbf a_u
\right),
& \text{Direct-Calibrated}.
\end{cases}
\end{equation}

where $\alpha=\sigma(\gamma)$ is a learned scalar. For nodes with no prior same-batch interaction, the reconstructed state defaults to the pre-batch anchor, such that when $\mathcal{H}_{u,k}=\emptyset$, all variants set
$\tilde{\mathbf h}_{u,k}=\mathbf a_u$.
\begin{table}[t]
\centering
\caption{Comparison of state reconstruction strategies
on Wikipedia ($B=4000$, $K=16$, $d_q=16$, $d_r=64$).
Speedup is measured relative to direct reconstruction.}
\label{tab:reconstruction_ablation}
\small
\setlength{\tabcolsep}{5pt}
\begin{tabular}{lcccc}
\toprule
Method & Val AP & Test AP & New AP & Speedup \\
\midrule
Residual
& \textbf{0.9766}
& \textbf{0.9707}
& \textbf{0.9664}
& $0.998\times$ \\
Direct
& 0.9702
& 0.9609
& 0.9617
& \textbf{$1.000\times$} \\
Direct-Calibrated
& 0.9701
& 0.9593
& 0.9593
& $0.993\times$ \\
\bottomrule
\end{tabular}
\end{table}

As shown in Table~\ref{tab:reconstruction_ablation}, residual reconstruction achieves the strongest predictive performance across the evaluated variants. Relative to direct reconstruction, it improves test AP from $0.9609$ to $0.9707$, while also improving new-node AP from $0.9617$
to $0.9664$. The direct-calibrated variant does not close this gap, indicating that simply
constraining a full-state prediction toward the anchor is insufficient. All three variants
exhibit nearly identical training speed, indicating
that the accuracy improvements of residual
reconstruction incur negligible computational
overhead. Intuitively, the pre-batch anchor already captures the node's established state, so the reconstruction network only needs to model the change induced by preceding same-batch interactions rather than relearning the full state.
These results support anchor-relative residual prediction, as preserving the known pre-batch anchor and learning only the state change induced by preceding same-batch interactions provide a more effective inductive bias than directly reconstructing the full state.

\section{conclusion}
\label{sec:conclusion}

We introduced AnchorRM, a causal intra-batch state reconstruction
mechanism for memory-based temporal graph neural networks. Instead of
reusing stale pre-batch memories for multiple interactions within the
same temporal batch, AnchorRM reconstructs each event-specific state
from a fixed pre-batch anchor and its causally preceding same-batch
history, while preserving parallel processing.
Experiments show that AnchorRM substantially reduces the accuracy
degradation caused by large temporal batches, with particularly strong
improvements for new-node prediction. These results indicate that
intra-batch state reconstruction enables more aggressive temporal batching
while retaining high predictive accuracy. Overall, our approach identifies
stale intra-batch memory as an important limitation to scaling memory-based
temporal graph models and shows that causal state reconstruction provides
an effective way to mitigate it.

\clearpage

\bibliographystyle{plain}
\bibliography{anchorrm_references}

\end{document}